\documentclass[letterpaper, 10 pt, journal, twoside]{IEEEtran}

\usepackage{amsmath} 
\usepackage{amssymb}  
\usepackage{biblatex} 
\usepackage{booktabs}
\usepackage{multirow}
\usepackage{graphicx}
\usepackage{subfigure}
\usepackage{tikz}
\usepackage{textcomp}
\usepackage{stfloats}
\usepackage{url}
\usepackage{verbatim}
\newcommand\red[1]{\textcolor{black}{{#1}}}

\begin{document}

\title{TransCG: A Large-Scale Real-World Dataset for Transparent Object Depth Completion and A Grasping Baseline
}

\author{Hongjie Fang$^1$, Hao-Shu Fang$^1$, Sheng Xu$^1$ and Cewu Lu$^2$
\thanks{Manuscript received: February, 21, 2022; Revised May, 6, 2022; Accepted June, 1, 2022.}
\thanks{This paper was recommended for publication by Editor Markus Vincze upon evaluation of the Associate Editor and Reviewers' comments.} 
\thanks{$^1$Hongjie Fang, Hao-Shu Fang and Sheng Xu are with the Department of Computer Science, Shanghai Jiao Tong University, Shanghai 200240, China (e-mail: galaxies@sjtu.edu.cn; fhaoshu@gmail.com; xs1020@sjtu.edu.cn).}
\thanks{$^2$Cewu Lu is the corresponding author, member of Qing Yuan Research Institute and MoE Key Lab of Artificial Intelligence, AI Institute, Shanghai Jiao Tong University, Shanghai 200240, China (e-mail: lucewu@sjtu.edu.cn).}
\thanks{Digital Object
Identifier (DOI): see top of this page.}
}

\markboth{IEEE Robotics and Automation Letters. Preprint Version. Accepted June, 2022}
{Fang \MakeLowercase{\textit{et al.}}: TransCG: A Large-Scale Real-World Dataset for Transparent Object Depth Completion and A Grasping Baseline} 

\maketitle

\begin{abstract}
Transparent objects are common in our daily life and frequently handled in the automated production line. Robust vision-based robotic grasping and manipulation for these objects would be beneficial for automation. However, the majority of current grasping algorithms would fail in this case since they heavily rely on the depth image, while ordinary depth sensors usually fail to produce accurate depth information for transparent objects owing to the reflection and refraction of light. In this work, we address this issue by contributing a large-scale real-world dataset for transparent object depth completion, which contains 57,715 RGB-D images from 130 different scenes. \red{Our dataset is the first large-scale, real-world dataset that provides ground truth depth, surface normals, transparent masks in diverse and cluttered scenes. Cross-domain experiments show that our dataset is more general and can enable better generalization ability for models.} Moreover, we propose an end-to-end depth completion network, which takes the RGB image and the inaccurate depth map as inputs and outputs a refined depth map. Experiments demonstrate superior efficacy, efficiency and robustness of our method over previous works, and it is able to process images of high resolutions under limited hardware resources. Real robot experiments show that our method can also be applied to novel transparent object grasping robustly. The full dataset and our method are publicly available at \url{www.graspnet.net/transcg}.
\end{abstract}


\section{Introduction}\label{sec:introduction}
Transparent materials are widely used in modern industry, and robots inevitably need to process transparent objects no matter in manufacturing, logistics, or household services. \red{Recently, much progress has been made in the field of robot grasping and manipulation~\cite{fang2020graspnet, mahler2019learning}. However, many of the advances are not directly applicable in scenes with transparent objects since these methods heavily rely on the depth information collected by the RGB-D cameras, yet ordinary depth sensors usually fail to construct a complete depth image in scenes that include transparent objects.} The physical properties of transparent objects would lead to the distortion of light path by reflection and refraction, resulting in noisy depth maps. Therefore, many depth-based algorithms are incapable to handle transparent objects such as plastic bottles and glass containers which can be found everywhere in our daily life.

The geometry estimation of transparent objects remains a challenging task in the computer vision field, though progress has been made by many researchers. Ba \textit{et al.}~\cite{ba2020deep} utilized a special polarization camera to leverage polarization cues for shape estimation and reached satisfactory results, while Li \textit{et al.}~\cite{li2020through} proposed a two-stage physical-based network to reconstruct the shape of the transparent objects using multi-view images and material prior. However, both methods require specialized hardware, which is not a general setting for robotic manipulation. \red{A more common setting is a robot arm with an RGB-D camera, which is the setting that we mainly focus on.}

\begin{figure}[t]
	\centering
	\includegraphics[width=0.48\textwidth]{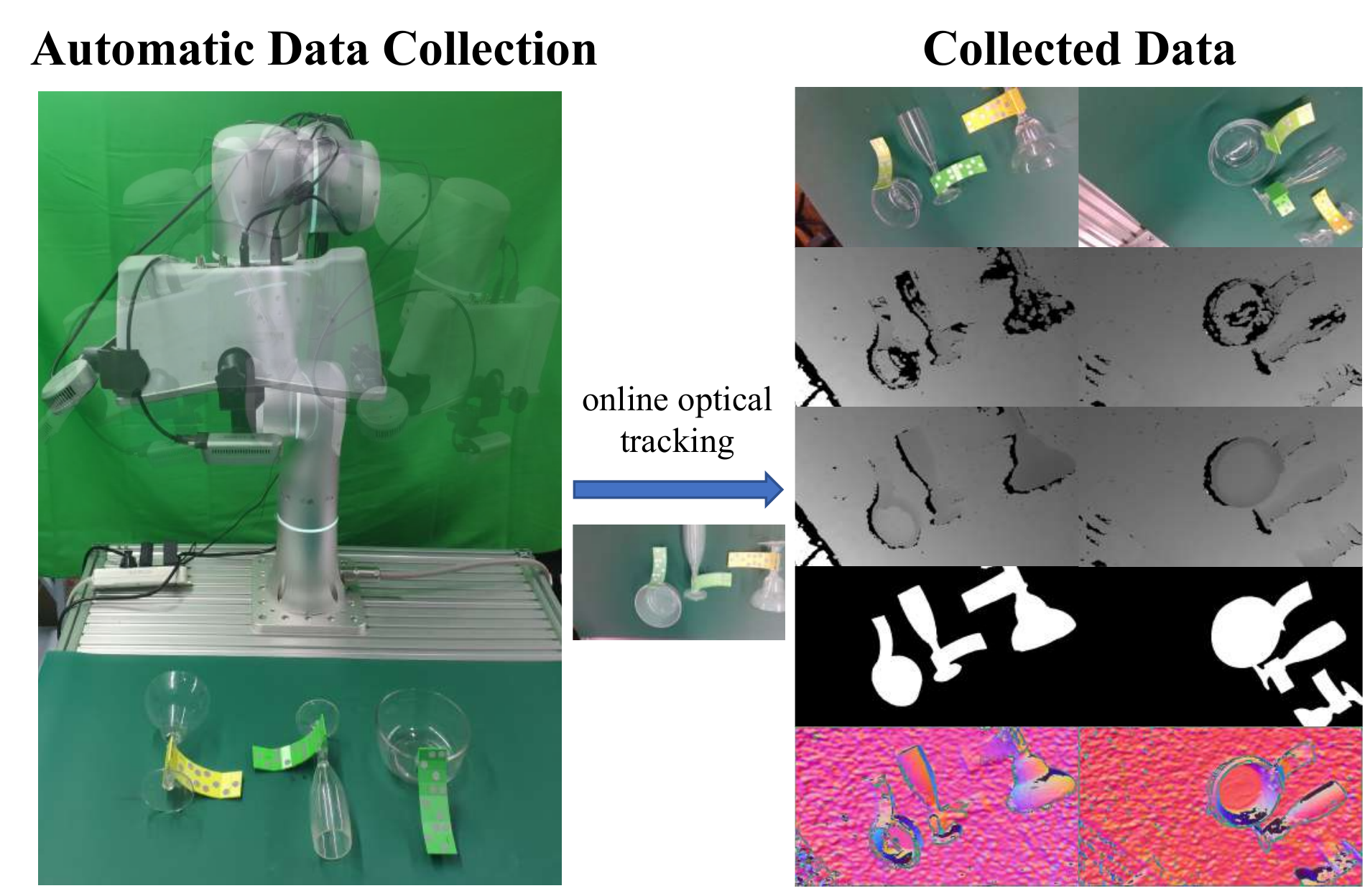}
	\caption{The methodology for building our dataset. We first perform object-level annotation to enable the real-time object pose tracking for our collecting system. With the assistance of online optical tracking, we are able to collect and annotate data automatically. Our dataset consists of RGB images, raw depth information collected by the depth sensor of RGB-D camera, the generated ground-truth depth images, transparent ground-truth mask and the surface normal from top to bottom. Our methodology can significantly reduce the workload of manual annotation and generate a large amount of high-quality real-world data. }
	\label{fig:methodology}
	\vspace{-0.4cm}
\end{figure}
Under this circumstance, Sajjan \textit{et al.}~\cite{sajjan2020clear} adapt the depth completion pipeline~\cite{zhang2018deep} to scenes that contain transparent objects and then propose ClearGrasp, which predicts the surface normal and the transparent boundary, followed by the global optimization to solve the depth estimation. A synthetic dataset and a small real-world dataset are also proposed along with the method. Zhu \textit{et al.}~\cite{zhu2021rgb} present an end-to-end framework for depth completion using the local implicit depth function, along with a synthetic Omniverse Object dataset. \red{Both synthetic datasets provide images containing transparent objects and their ground-truth depth maps, but the lack of real depth maps degrades the performance of those methods \cite{sajjan2020clear, zhu2021rgb} in real-world applications inevitably.}

\red{To close the synthetic-to-real gap in the field of grasping concerning transparent objects, we propose \textbf{TransCG}, a large-scale real-world dataset for transparent object depth completion.} A novel semi-automatic pipeline is proposed to accelerate the data collection and annotation process. \red{In total, our dataset contains \textbf{57,715 RGB-D images} of \textbf{51 transparent objects} and around 200 opaque objects captured from different perspectives of \textbf{130 scenes} under real-world settings.} The 3D mesh model of the transparent objects are also provided in our dataset. The methodology for building our dataset is shown in Fig. \ref{fig:methodology}.

\red{Futhermore, we propose a robust, efficient and effective network \textit{Depth Filler Net} (\textbf{DFNet}) for depth completion based on our dataset, which allows it to assist 6-DoF grasping methods on transparent object grasping. The quantitative and qualitative results show that our proposed DFNet (1) generalizes better across different scenes compared to existing methods; (2) improves most of the metrics to a great extent; (3) yields the highest inference speed and consumes the least computation overhead. We also apply our network to real-world object grasping for novel transparent objects, and a promising performance is witnessed. The full dataset, source code and pretrained models are released at \url{www.graspnet.net/transcg}}.

\section{Related Works}\label{sec:relatedwork}

\subsection{Transparent Datasets}\label{subsec:transdataset}

Due to the special optical properties of transparent objects which leads to undetermined and inaccurate results of optical sensors, the datasets concerning of transparent objects is usually difficult to build. 

\red{For tasks that do not need the depth information, \textit{e.g.}, transparent object classification, segmentation and pose estimation, there are large-scale datasets such as Trans10K-V2~\cite{xie2021segmenting}, Stanford2D-3D~\cite{armeni2017joint} and StereOBJ-1M~\cite{liu2021stereobj}.} 
For datasets that needs accurate sensor information as ground-truth, the common solutions are synthetic datasets, such as ClearGrasp synthetic dataset~\cite{sajjan2020clear} and Omniverse object dataset~\cite{zhu2021rgb}. \red{Built by tools like SuperCaustics~\cite{mousavi2021supercaustics}, the greatest shortcoming to those datasets is that the raw information collected by sensors is unlikely to be obtained in simulation}.
\red{Xu \textit{et al.} \cite{xu2022seeing} propose Toronto Transparent Object Depth Dataset (TODD), another real-world depth completion dataset. However, the diversities of transparent objects, translucent objects and cameras in the dataset are limited.}
Liu \textit{et al.} 
\cite{liu2020keypose} build a real-world keypoint estimation dataset for transparent objects, which consists of the ground-truth depth map generated by substituting the transparent object with the identical opaque objects. But the large amount of real-world samples are all captured alone under certain environments, which is rare in real-world applications. 
The dataset collection approach is also used in ClearGrasp real-world dataset~\cite{sajjan2020clear} which only has 286 samples since generating the missing information is both time-consuming and labor-consuming. 

\red{To overcome the difficulties of building datasets concerning transparent objects, we propose a novel pipeline for transparent dataset construction. Using the pipeline, we build TransCG, a complete large-scale real-world dataset for transparent object depth completion. Details will be introduced in Section \ref{sec:dataset}.}

\subsection{Depth Completion and Estimation}\label{subsec:depthestimation}

\red{Depth completion and estimation has been studied by many researchers for a long time, and can be categorized into three classes: estimating depth directly from an RGB image~\cite{eigen2014depth, garg2016unsupervised, klingner2020self, yin2018geonet}, estimating depth from an RGB image with sparse depth information~\cite{chen2018estimating, cheng2018depth, fu2020depth, lee2021depth, ma2018sparse} and depth completion from an RGB image with inaccurate depth information~\cite{sajjan2020clear, zhang2018deep, xu2022seeing, zhu2021rgb}. Since depth information concerning reflective objects is usually noisy and inaccurate, transparent object depth completion falls into the last category, and it can be further classified into multi-view depth completion and single-view depth completion.}

\red{For multi-view depth completion, Ichnowski \textit{et al.} \cite{ichnowski2021dex} use NeRF to recover depth for grasping transparent objects. 
Li \textit{et al.} \cite{li2020through} present a physical-based network, which uses multi-view images to recover high-quality 3D geometry of transparent objects.}

\def\halfcheckmark{\tikz\draw[scale=0.4,fill=black](0,.35) -- (.25,0) -- (0.8,.6) -- (.25,.15) -- cycle (0.7,0.2) -- (0.8,0.2)  -- (0.4,0.5) -- cycle;}

\def\halfredcheckmark{\tikz\draw[scale=0.4,fill=black](0,.35) -- (.25,0) -- (0.8,.6) -- (.25,.15) -- cycle (0.7,0.2) -- (0.8,0.2)  -- (0.4,0.5) -- cycle;}

\def\fullcheckmark{\tikz\draw[scale=0.4,fill=black](0,.35) -- (.25,0) -- (0.8,.6) -- (.25,.15);}

\def\fullredcheckmark{\tikz\draw[scale=0.4,fill=black](0,.35) -- (.25,0) -- (0.8,.6) -- (.25,.15);}

\begin{table*}
    \vspace{5pt}
    \setlength\tabcolsep{3pt}
    \centering
    \begin{minipage}[t]{0.95\linewidth}
    \begin{center}
        \caption{Comparisons of Transparent Depth Completion Datasets}\label{tab:dataset}
        \begin{tabular}{|c|c|c|c|c|c|c|c|c|c|c|c|c|c|}
            \hline
            \multirow{2}{*}{\textbf{Type}} & \multirow{2}{*}{\textbf{Dataset}} &
            \multicolumn{3}{c|}{\textbf{Dataset Completeness}} & \multicolumn{3}{c|}{\textbf{Scene Type}} & \multirow{2}{*}{\red{\textbf{Auto-Collection}}} & \multirow{2}{*}{\textbf{\#Cam.}} & \multirow{2}{*}{\textbf{\#Obj.}} & \multirow{2}{*}{\textbf{\#Img.}} \\ \cline{3-8}
            & & RGB & Raw depth & Refined depth & Single & Isolated & Cluttered & & & & \\ \hline
            \multirow{2}{*}{Syn} & Clear-Syn \cite{sajjan2020clear} & \fullcheckmark & & \fullcheckmark & \fullcheckmark & \fullcheckmark & & \red{\fullredcheckmark} & - & 9 & 50K \\
            & OOD \cite{zhu2021rgb} & \fullcheckmark & & \fullcheckmark & \fullcheckmark & \fullcheckmark & \fullcheckmark & \red{\fullredcheckmark} & - & 9 & 60K \\
            \hline
            \multirow{4}{*}{Real} & Clear-Real \cite{sajjan2020clear} & \fullcheckmark & \fullcheckmark & \fullcheckmark & \fullcheckmark & \fullcheckmark & & & 2 & 10 & 286 \\
            & TOD \cite{liu2020keypose} & \fullcheckmark & \fullcheckmark & \fullcheckmark & \fullcheckmark & & & & 1 & 15 & 48K \\
            & \red{TODD \cite{xu2022seeing}} & \red{\fullredcheckmark} & \red{\fullredcheckmark} & \red{\fullredcheckmark}  &\red{\fullredcheckmark} & \red{\fullredcheckmark} & 
             & \red{\fullredcheckmark} & \red{1} & \red{6} & \red{15K} \\
            & TransCG (ours) & \fullcheckmark & \fullcheckmark & \fullcheckmark  &\fullcheckmark & \fullcheckmark & \fullcheckmark & \red{\fullredcheckmark} & 2 & 51 & 58K \\
            \hline
        \end{tabular}
        \end{center}
        \vspace{-0.15cm}
        \footnotesize{\textbf{Note.} Clear refers to ClearGrasp Dataset (Synthetic/Real-world), OOD refers to Omniverse Object Dataset, TOD refers to Transparent Object Dataset. ``\#Cam.'' denotes types of RGB-D cameras, ``\#Obj.'' stands for the number of target objects and ``\#Img.'' represents the amount of samples. \red{Here cluttered scenes refer to scenes that consists of more than 3 transparent objects and several opaque objects.}}
    \end{minipage}
    \vspace{-0.5cm}
\end{table*}

\red{For single-view depth completion, Zhang \textit{et al.}~\cite{zhang2018deep} propose a two-stage depth completion pipeline, which predicts surface normals and occlusion boundaries according to the RGB image, followed by the global optimization to complete the depth map.} 
ClearGrasp~\cite{sajjan2020clear} makes a few critical modifications to the pipeline and adapts it in depth completion concerning transparent object, but its inference speed is unacceptable in real-time grasping scenarios. \red{Tang \textit{et al.} \cite{tang2021depthgrasp} utilize self-attentive adversarial network to replace the global optimization in ClearGrasp.} 
Zhu \textit{et al.}~\cite{zhu2021rgb} propose a two-stage system consisting of local implicit depth function prediction and depth refinement. Though it outperforms previous works on speed and accuracy, its generalization ability is very limited according to our cross-domain tests in Sec.~\ref{sec:cross_domain}, which makes it difficult to fit into real-world robotic manipulation settings.
\red{A concurrent work \cite{xu2022seeing} combines previous depth completion method \cite{senushkin2020decoder} with point cloud completion to improve the quality of the refined depth.}

To achieve a better generalization and applicability in real-world environment, a tiny but robust model is needed, which requires a large amount of real-world data as support.

\subsection{6-DoF Grasping}\label{subsec:6dofgrasping}
6-DoF grasping refers to those methods that predict the position and rotation of the gripper in 3D domain. Enabling the robots to grasp objects from various angles, it is the basis of robotic manipulation. Due to the versatility and effectiveness of 6-DoF grasping, it is currently the main stream in the grasping field.

GPD~\cite{ten2017grasp} propose a two-stage 6-DoF grasping method, which estimates grasp candidates sampled under empirical constraints. 
PointNetGPD~\cite{liang2019pointnetgpd} improves GPD by adapting PointNet~\cite{qi2017pointnet} in evaluation. 
Mousavian \textit{et al.}~\cite{mousavian20196} leverage  variational auto-encoder to sample grasp poses, and add refinement process after evaluation for better performance. \red{Qin \textit{et al.}~\cite{qin2020s4g} regress the grasp pose directly from the partial-view point cloud,} while 
Ni \textit{et al.}~\cite{ni2020pointnet++} regress the grasp pose from features extracted by PointNet++~\cite{qi2017pointnet++}.
Fang \textit{et al.}~\cite{fang2020graspnet} propose the GraspNet-1Billion dataset for general object grasping and an end-to-end grasp pose prediction network. 
Gou \textit{et al.}~\cite{gou2021rgb} incorporate RGB and depth information to improve the performance of 6-DoF grasping.
\red{Sundermeyer \textit{et al.}~\cite{sundermeyer2021contact} reduce 6-DoF grasp poses to 4-DoF grasp representations on a known contact point to facilitate the learning problems and improve the grasping quality in cluttered scenes.}
\red{Wang \textit{et al.}~\cite{wang2021graspness} propose a geometrically-based quality graspness to evaluate the graspable area in cluttered scenes.}

All these methods rely heavily on depth image, which makes them unsuitable for transparent object grasping. Thus, utilizing color information to generate high-quality depth map and point cloud to aid the grasping deserves further exploration.

\section{Dataset}\label{sec:dataset}

\subsection{Overview}\label{subsec:overview}

\begin{figure}[b]
	\centering
	\includegraphics[width=0.48\textwidth]{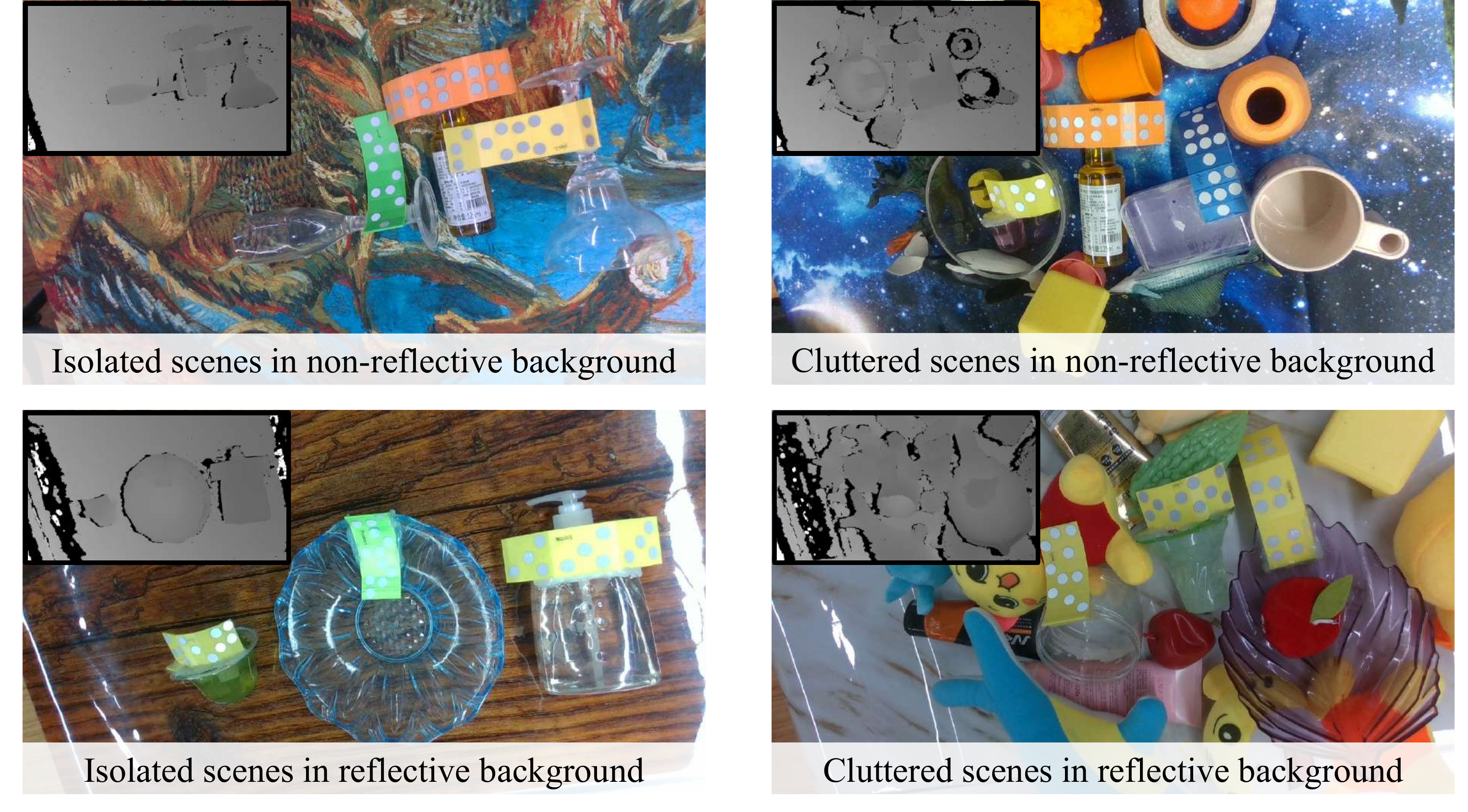}
	\caption{Different types of scenes in our TransCG dataset. The sub-figures on the top left are the refined ground-truth depth images.} 
	\label{fig:scenes}
\end{figure}

\begin{figure*}
    \vspace{5pt}
	\centering
	\includegraphics[width=0.9\textwidth]{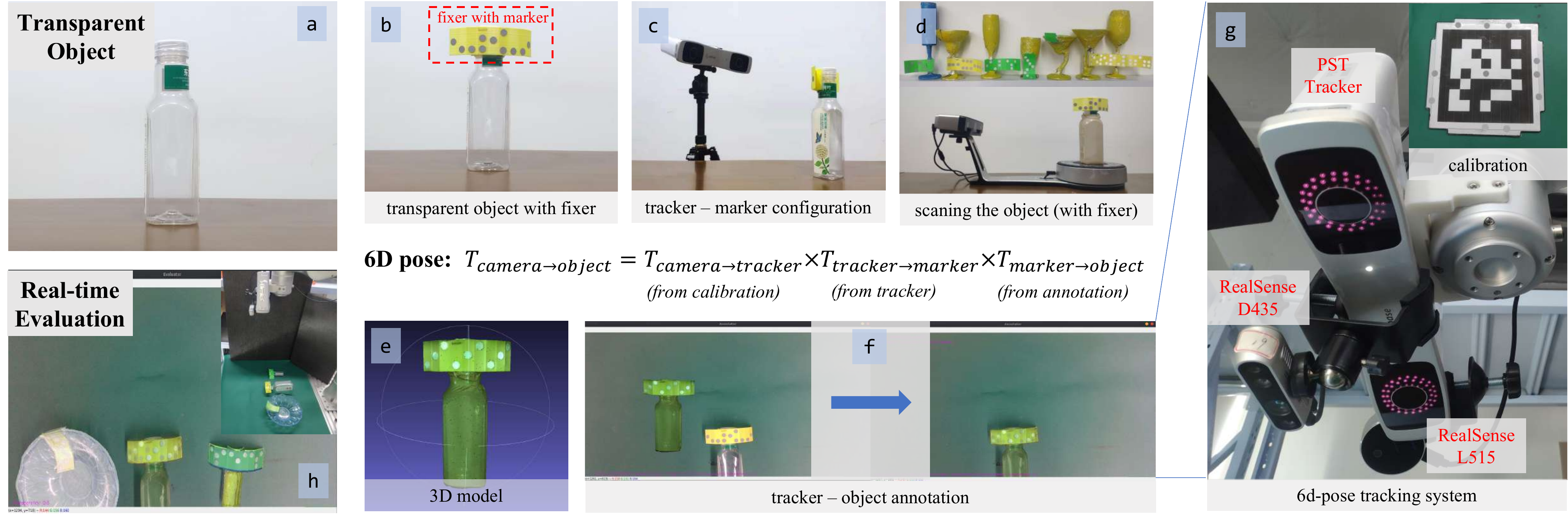}
	\caption{System setup process. Given a transparent object (a), we attach a fixer with IR markers to it (b) and record its pattern with an optical tracker (c), which can enable tracking afterwards. Then we scan the object (d) and get its 3D model (e). After that, we manually perform tracker-object annotation to get the transformation matrix from marker to object (f), where an GUI is developed for real-time evaluation (h). The whole annotation and tracking process is assisted by our 6D pose tracking system (g), which consists of a PST tracker, an Intel RealSense D435 camera and an Intel RealSense L515 camera.}
	\label{fig:annotation}
	\vspace{-0.5cm}
\end{figure*}

As introduced before, the previous transparent datasets that need accurate ground-truth depth usually use simulation platforms to generate synthetic data, and real-world datasets are usually small in scale due to the workload of annotations. To overcome the difficulties, we propose a novel pipeline to build the dataset efficiently, which utilizes a robot to perform data collection after limited object-level manual annotations. 

\red{In brief, we aim to collect a dataset that contains RGBD images using real-world sensors, along with detail annotations for the transparent objects include their depths, masks, 6D poses, normals, etc.} To reduce the annotation effort, our methodology for building the dataset is to automatically localize the transparent objects during data collection. To achieve that, we resort to an optical tracker that can accurately localize a target's 6D pose in real-time from several IR markers attached to it. Besides, we manage to obtain the 3D model of each transparent object in the training set by wrapping them with opaque materials and scanning with a 3D scanner. With these two preliminaries, we can easily obtain the transparent objects' 6D pose during data collection. The geometry of transparent objects can be restored by leveraging the tracker results and their 3D models during data annotation. Our pipeline is able to build large-scale real-world dataset conveniently and reduce the human workload by a large extent.

Using the pipeline, we build our TransCG dataset, which contains 57,715 RGB-D images captured by two different cameras, along with the refined ground-truth depth images, transparent ground-truth mask and the surface normals, from 130 scenes under various background settings, within a week. We collect 51 common objects in daily life that can lead to inaccurate depth map, including transparent objects, translucent objects, reflective objects and objects with dense tiny holes. Apart from 65 simple isolated scenes that are similar to the scenes in the previous datasets, we also provide 65 challenging cluttered scenes that are closer to the real-world grasping environment, as shown in Fig. \ref{fig:scenes}. More details are presented in supplementary video. The comparisons between our dataset and existing datasets are summarised in Tab. \ref{tab:dataset}.

\subsection{System Setup}\label{subsec:dataannotation}

To support fast and accurate data collection process, we build a transparent object tracking system, which is the only part that requires human efforts in our dataset building pipeline. The system setup process is illustrated in Fig. \ref{fig:annotation}.

\red{Given a transparent object, we firstly attach a fixer with IR markers to it, as shown in Fig. \ref{fig:annotation}b. A PST optical tracker\footnote{\url{https://www.ps-tech.com/products-pst-base/}} is used to record the IR markers and tracks them afterwards. Although we can directly attach IR markers to the object, we found that using a flat fixer can make the tracking more robust. Next, we temporarily wrap the transparent object with some opaque materials that can preserve the shape of objects, and obtain its 3D model with a Shining3D EinScan-SP scanner\footnote{\url{https://www.einscan.com/desktop-3d-scanners/einscan-sp/}}. With these two steps, we can obtain the transparent object's 6D pose during data collection, where the details is further explained.
}

The core of our data collection system consists of a PST optical tracker, an Intel RealSense D435 camera and an Intel RealSense L515 camera. The tracker outputs the 6D pose of the markers in real-time and the two cameras provide RGBD images with different quality. The 6D pose of the $i$-th object \textit{w.r.t} the $j$-th camera $\mathbf{T}_{\mathrm{cam}_j}^{\mathrm{obj}_i}$ can be calculated as follows:
\begin{equation}
\mathbf{T}_{\mathrm{cam}_j}^{\mathrm{obj}_i} = \mathbf{T}_{\mathrm{cam}_j}^{\mathrm{tracker}}\mathbf{T}_{\mathrm{tracker}}^{\mathrm{marker}_i}\mathbf{T}_{\mathrm{marker}_i}^{\mathrm{obj}_i},
\end{equation}

\noindent where $\mathbf{T}_{\mathrm{cam}_j}^{\mathrm{tracker}}$ denotes the transformation matrix of the tracker origin \textit{w.r.t} the $j$-th camera,  $\mathbf{T}_{\mathrm{tracker}}^{\mathrm{marker}_i}$ denotes the 6D pose of the markers attached on the $i$-th object \textit{w.r.t} the tracker, and $\mathbf{T}_{\mathrm{marker}_i}^{\mathrm{obj}_i}$ denotes the transformation matrix of the $i$-th object's origin \textit{w.r.t} the markers. 

To obtain $\mathbf{T}_{\mathrm{cam}_j}^{\mathrm{tracker}}$, we perform tracker-camera calibration using a calibration board with both Aruco marker and IR markers.  $\mathbf{T}_{\mathrm{tracker}}^{\mathrm{marker}_i}$ is the output of the optical tracker, and $\mathbf{T}_{\mathrm{marker}_i}^{\mathrm{obj}_i}$ 
are annotated by human. We develop a GUI application for the annotation process and evaluate the results in real time by rendering the objects in the corresponding RGB image.


\red{On average, the overall human efforts to process an object is around 1 hour.} With the transparent object tracking system we introduced, we can easily recover the ground-truth depth information of the transparent objects afterwards.

\subsection{Data Collection}\label{subsec:datacollection}

To collect a large amount of data automatically, we attach our tracking system to a robot arm that moves along a fixed trajectory containing 240 distinct viewpoints. Transparent objects are randomly selected and placed in the scene. To align with real-world setting, we augment the scene with various table covers and opaque objects. At each viewpoint, the tracking system will capture the RGBD images and the tracking results.

Before data collection, we perform camera extrinsic calibration for the 240 viewpoints, which can assist to recover the 6D pose of objects in case of tracker failure. ArUco markers are used during camera calibration to obtain the pose of the $j$-th camera $\mathbf{T}^{\mathrm{cam}_j}_k$ \textit{w.r.t} the marker coordinate in the $k$-th viewpoint for all $k$. Specifically, the 6D pose of the $i$-th object \textit{w.r.t} the $j$-th camera in the perspective $k$ can be obtained from the corresponding 6D pose in perspective $k_0$ and the camera poses in viewpoints $k, k_0$, \red{where $k_0$ is the successfully-tracked viewpoint}:
\begin{equation}\label{eqn:cam}
\left(\mathbf{T}_{\mathrm{cam}_j}^{\mathrm{obj}_i}\right)_{k} = \mathbf{T}^{\mathrm{cam}_j}_k \left(\mathbf{T}^{\mathrm{cam}_j}_{k_0}\right)^{-1}\left(\mathbf{T}_{\mathrm{cam}_j}^{\mathrm{obj}_i}\right)_{k_0}.
\end{equation}

Thus, we can recover the 6D pose of the objects that are not tracked in some viewpoints using the tracking results in the first viewpoint, which are always successfully detected.

After collecting the raw data, we use the collected 6D poses to render the ground-truth depth maps, the transparent masks and ground-truth surface normals.

\subsection{Dataset Postprocessing}\label{subsec:datasetsplit}
\subsubsection{Data Verification} 

\red{The blurry samples and improperly exposed samples are automatically detected and removed from the dataset using tools like Laplacian operators and histograms provided in OpenCV library \cite{bradski2000opencv}. Manual validations are conducted by rendering the object mesh to the scene using the 6D pose and see if they can overlap with the original object.} Finally, we generate a metadata containing all valid viewpoints for each scene.

\subsubsection{Dataset Split} 

\red{We randomly select 12 objects from different categories and regard all scenes that contain these objects as the testing set}, while the other scenes are used as the training set. There are totally 34,191 training samples and 23,524 testing samples.

\subsection{Discussion}\label{subsec:discussion}

\begin{figure*}
\vspace{5pt}
	\centering
	\includegraphics[width=0.9\textwidth]{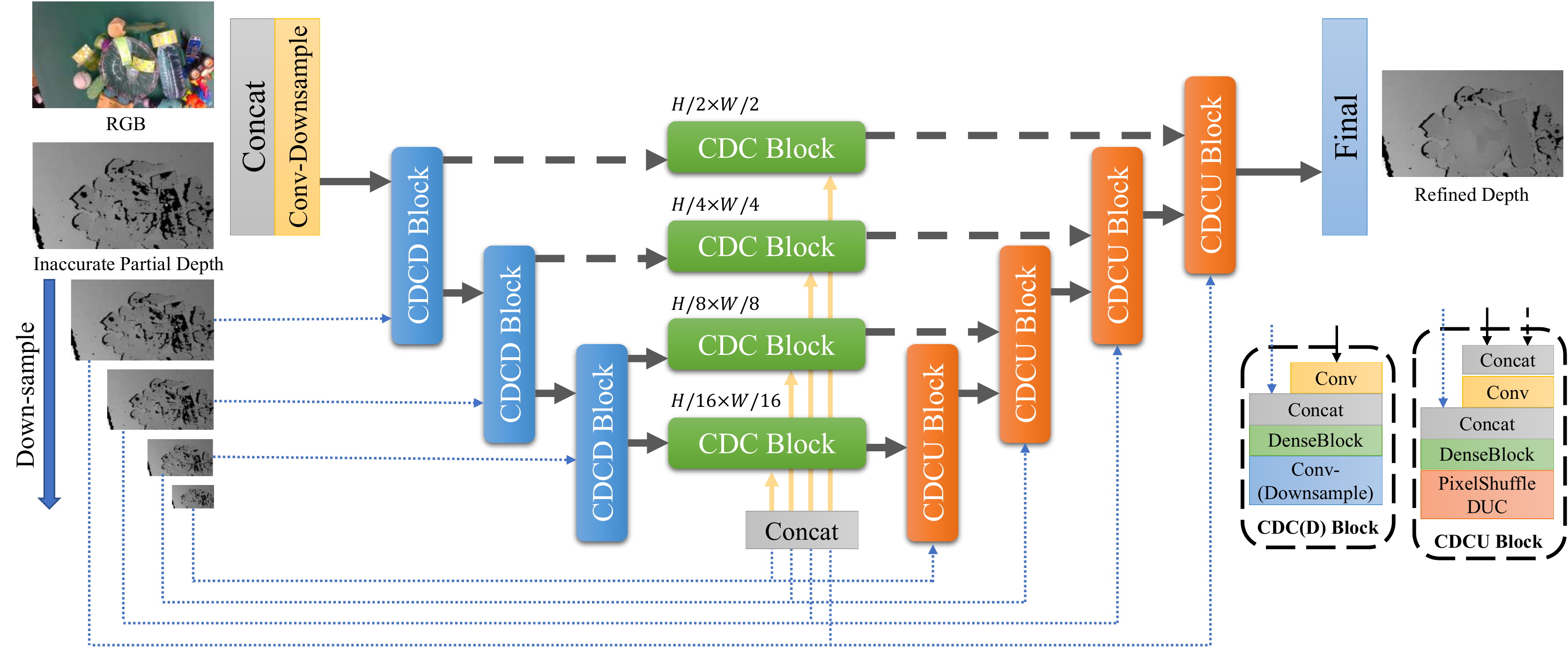}
	\caption{The architecture of our proposed end-to-end depth completion network DFNet. Our network utilizes a U-Net architecture with CDCD blocks, CDC blocks and CDCU blocks. These blocks are mainly composed of dense blocks~\cite{huang2017densely}, with DUC~\cite{wang2018understanding} replacing deconvolution layer in up-sampling of CDCU block. All convolution layers except the last one are followed by batch normalizations~\cite{ioffe2015batch} and ReLU activations, and have $3\times 3$ kernels.}
	\label{fig:network}
	\vspace{-0.5cm}
\end{figure*}
\red{To ensure that the markers have minimal influence to the object, we do not directly place them on the surface of transparent objects. Instead they are placed on the surface of a fixer that is attached to the object. Therefore, we can make sure that the transparent part of the object is retained to the greatest extent. From the cross-dataset experiment conducted in Table~\ref{tab:cross_domain} and Table~\ref{tab:cross_domain_dataset}, we can see that methods trained on our dataset generalize well to other testing datasets. These experiments demonstrate that the markers have minimal impact on the training of network.}
\section{Method}\label{sec:method}

\subsection{Overview}\label{subsec:methodoverview}
In this section, we detail our method for depth completion and grasping. For depth completion, we propose an end-to-end network which is illustrated in Fig. \ref{fig:network}. Given an RGB image $\mathcal{C}\in\mathbb{R}^{H\times W\times 3}$ and an inaccurate partial depth image $\mathcal{D}\in\mathbb{R}^{H\times W}$, our network completes the depth information and predicts the full depth map $\hat{\mathcal{D}} \in \mathbb{R}^{H\times W}$, where $H \times W$ is the size of an image. The details of the network will be introduced in Sec. \ref{subsec:depthcompletion}. In Sec. \ref{subsec:objectgrasping}, \red{we demonstrate how our depth completion network can be applied to the grasping task and serve as a grasping baseline method.} we apply our depth completion to a grasp pose detection network~\cite{fang2020graspnet} which takes point cloud as inputs and outputs the grasp poses. We demonstrate that our network can generate high quality depth for transparent objects that can enable depth-based grasping.

\subsection{Depth Completion}\label{subsec:depthcompletion}

Inspired by previous literature~\cite{chen2018estimating} about depth estimation from sparse sensing, we propose our end-to-end depth completion network \textit{Depth Filler Net} (\textbf{DFNet}) that predicts full depth map according to RGB information and inaccurate partial depth. Adapting a U-Net architecture with depth of four layers, our network takes dense blocks~\cite{huang2017densely} as backbones and formulates \textit{Conv-Dense-Conv-Downsample} (CDCD) blocks, \textit{Conv-Dense-Conv} (CDC) blocks and \textit{Conv-Dense-Conv-Upsample} (CDCU) blocks. Empirical statistics about depth estimations and completions show that original depth information is critical throughout the networks. Hence we provide the original depth information as an input to every CDCD, CDC and CDCU block. Skip paths are added to retain information in high resolutions. Inspired by AlphaPose~\cite{fang2017rmpe}, we adapt dense up-sampling convolution (DUC)~\cite{wang2018understanding} instead of ordinary deconvolution layers in CDCU blocks. 

Our network is trained using the following loss function:
\begin{equation}
    \mathcal{L} = \mathcal{L}_d + \beta \mathcal{L}_s,
\end{equation}

\noindent where $\beta$ is the weight parameter, $\mathcal{L}_d$ penalizes depth inaccuracy and $\mathcal{L}_{s}$ is the cosine distance of surface normals computed from predicted depth map and ground-truth depth map \cite{zhu2021rgb}, which penalizes unsmoothness. Formally,
\begin{equation}
    \begin{aligned}
        & \mathcal{L}_d = \left\|\hat{\mathcal{D}} - \mathcal{D}^*\right\|^2, \\
        & \mathcal{L}_s = 1 - \cos \left< \hat{\mathcal{D}}_h \times \hat{\mathcal{D}}_w, \mathcal{D}^*_h \times \mathcal{D}^*_w \right>,
    \end{aligned}
\end{equation}

\noindent where $\hat{\mathcal{D}}$ and $\mathcal{D}^*$ denotes the predicted depth and the ground-truth depth, $\mathcal{D}_w$ and $\mathcal{D}_h$ are gradient vectors along width-axis and height-axis of depth map $\mathcal{D}$ respectively. For both losses, we regard depths out of range $[0.3, 1.5]$ as invalid pixels and remove them from losses to reduce the impact of outliers.


\subsection{Object Grasping}\label{subsec:objectgrasping}

Given an RGB image along with a depth image collected by an RGB-D camera, we first scale the images to an appropriate size and feed into our depth completion model \textbf{DFNet}, which outputs the refined depth in the same resolution as the input. Then, the refined depth is scaled back to the original size, which can be used to construct the scene point cloud using camera intrinsics. After that, the scene point cloud is sent to GraspNet-baseline~\cite{fang2020graspnet} as the input to the end-to-end grasp pose detection network. \red{Other depth based grasping methods are also applicable.} Finally, the grasp pose detection network outputs the grasp candidates, and the grasp will be executed by a parallel-jaw robot.

\section{Experiments}\label{sec:experiments}

\subsection{Depth Completion Experiments}

\begin{table*}
\vspace{5pt}
    \setlength\tabcolsep{3pt}
    \centering
    \begin{minipage}[t]{0.76\linewidth}
    \begin{center}
    \caption{Experiment Results on TransCG Dataset}
    \label{tab:result}
    \begin{tabular}{|c|cccccc|c|c|c|}
    \hline
    \multirow{2}{*}{\textbf{Methods}} & \multicolumn{6}{c|}{\textbf{Metrics}} & \multirow{2}{*}{\begin{tabular}{c}\textbf{GPU Memory}\\ \textbf{Occupation}\end{tabular}} & \multirow{2}{*}{\begin{tabular}{c}\textbf{Inference}\\ \textbf{Time}\end{tabular}} & \multirow{2}{*}{\begin{tabular}{c}\textbf{Model}\\ \textbf{Size}\end{tabular}} \\ \cline{2-7} 
    &  RMSE $\downarrow$ & REL $\downarrow$ & MAE $\downarrow$ & $\delta_{1.05}$ $\uparrow$ & $\delta_{1.10}$ $\uparrow$ & $\delta_{1.25}$ $\uparrow$ & & & \\ \hline
    ClearGrasp \cite{sajjan2020clear} & 0.054 & 0.083 & 0.037 & 50.48 & 68.68 & 95.28 & 2.1 GB & 2.2813s & 934 MB \\
    LIDF-Refine \cite{zhu2021rgb} & 0.019 & 0.034 & 0.015 & 78.22 & 94.26 & \textbf{99.80} & 6.2 GB & 0.0182s & 251 MB \\
    \red{TranspareNet \cite{xu2022seeing}} & \red{0.026} & \red{\textbf{0.023}} & \red{0.013} & \red{\textbf{88.45}} & \red{\textbf{96.25}} & \red{99.42} & \red{1.9 GB} & \red{0.0354s} & \red{336 MB}\\
    DFNet (ours) & \textbf{0.018} & 0.027 & \textbf{0.012} & 83.76 & 95.67 & 99.71 & \textbf{1.6 GB} & \textbf{0.0166s} & \textbf{5.2 MB}\\ \hline 
    \end{tabular}
    \end{center}
    \vspace{-0.15cm}
    {\footnotesize \textbf{Note}. $\downarrow$ means lower is better, and $\uparrow$ means higher is better. 
    GPU Memory occupation and inference time are measured with an NVIDIA GeForce RTX 3090 GPU.}
    \end{minipage}
    \vspace{-0.1cm}
\end{table*}
\begin{figure*}
    \vspace{-0.3cm}
	\centering
	\includegraphics[width=\textwidth]{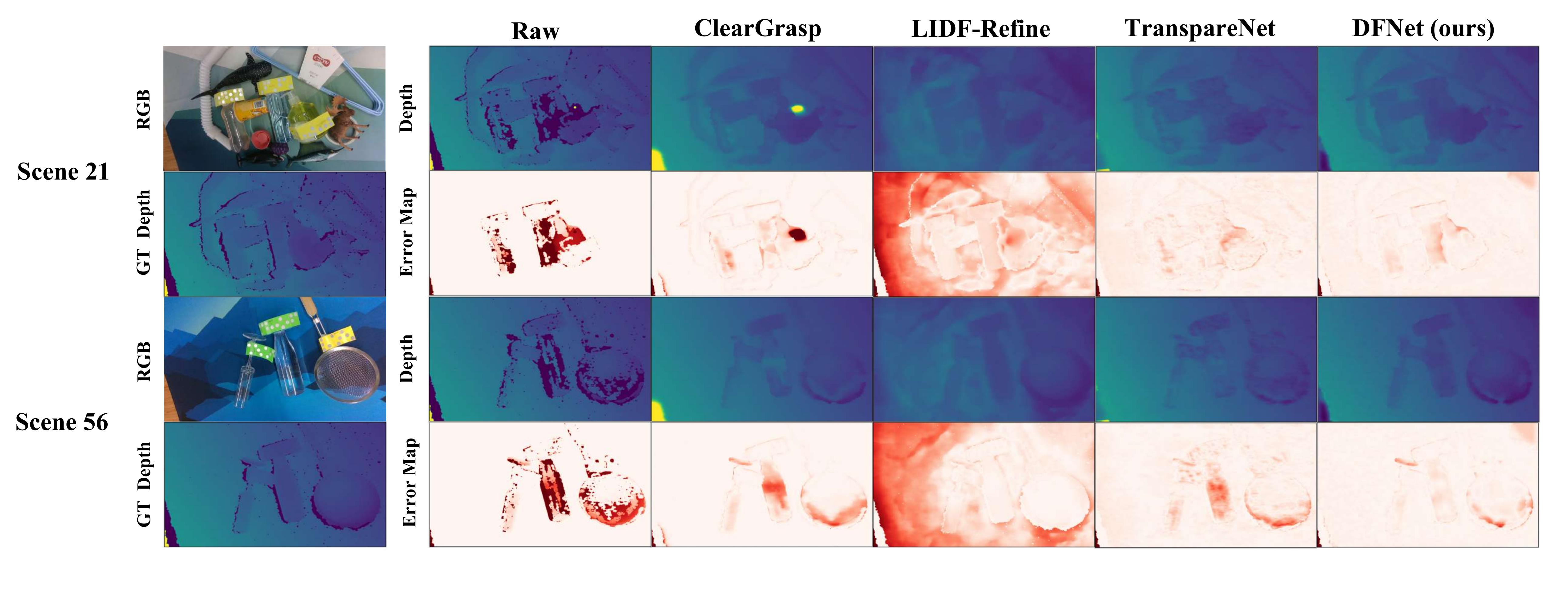}
	\vspace{-1cm}\caption{Visualizations of the refined depth maps and error maps of different methods on several testing scenes of TransCG dataset.} 
	\label{fig:expr-vis}
	\vspace{-0.5cm}
\end{figure*}
We compare our method with several representative approaches on our TransCG dataset. \red{ClearGrasp~\cite{sajjan2020clear} is the first algorithm which leverages deep learning with synthetic training data to estimate depth information concerning transparent objects, and LIDF-Refine~\cite{zhu2021rgb} utilizes local implicit depth function to solve transparent object depth completion task. We also evaluate the concurrent work TranspareNet \cite{xu2022seeing} in experiments.} All baselines are trained in our TransCG dataset using their released source codes and optimal hyper-parameters for fair comparisons.

For our model, the hidden channels in the network is set to 64. In every dense block, the layers $L$ and the feature channels of each layer $k$ are set to 5, 12 respectively, as suggested in literature~\cite{chen2018estimating}. We use AdamW optimizer~\cite{loshchilov2017fixing} with initial learning rate of $10^{-3}$ and multi-step learning rate scheduler which decays the learning rate by 5 after 5, 15, 25, 35 epochs. We train the model for 40 epochs with the batch size of 32. Several data augmentation approaches such as random flipping, rotation, noise adding and chromatic transformations in HLS color space are conducted during training. Concerning loss, we set $\beta = 0.001$.

For all methods, we scale the images to $320 \times 240$ during training and testing.
\red{We use 4 NVIDIA GeForce RTX 3090 GPUs for training and one for testing, and the average time to train an epoch is approximately 20 minutes.}

The following common metrics of depth completion for transparent objects are used in comparisons. All metrics are calculated on the transparent areas according to transparent masks unless specified.
\begin{itemize}
    \item RMSE: the rooted mean squared error between depth estimates and ground-truth depths.
    \item REL: the mean absolute relative difference.
    \item MAE: the mean absolute error between depth estimates and ground-truth depths.
    \item Threshold $\delta$: the percentage of pixels with predicted depths satisfying $\max(d/d^*, d^*/d) < \delta$, where $d, d^*$ are corresponding pixels of $\mathcal{D}, \mathcal{D}^*$, and $\delta$ is set to 1.05, 1.10 and 1.25 following~\cite{sajjan2020clear, zhu2021rgb}.
\end{itemize}

\red{The quantitative results are reported in Tab. \ref{tab:result}, and the qualitative results are visualized in Fig. \ref{fig:expr-vis}. Our method outperforms ClearGrasp \cite{sajjan2020clear} and LIDF-Refine \cite{zhu2021rgb} on both metrics and quality, and performs better than the concurrent work TranspareNet \cite{xu2022seeing} on some metrics. Qualitatively, it completes the inaccurate region of depth maps more completely (Fig. \ref{fig:expr-vis}) than \cite{xu2022seeing}. It is also worth noticing that though LIDF-Refine \cite{zhu2021rgb} performs well on masked metrics, it may introduce shadows on areas without transparent objects, which may result from the outliers of the original depth map. Moreover, our method has the smallest size, fastest inference time and lowest GPU memory occupation, which allows it to perform depth completion of high quality under limited resources (Tab. \ref{tab:result}).}

\subsection{Cross-Domain Experiments}
\label{sec:cross_domain}
\begin{table}[t]
    \vspace{0.3cm}
    \setlength\tabcolsep{1.5pt}
    \centering
    \begin{minipage}[t]{0.95\linewidth}
    \begin{center}
    \caption{Results of Cross-Domain Experiments of Methods}
    \label{tab:cross_domain}
    \begin{tabular}{|c|c|cccccc|} \hline
    \textbf{Training/} & \multirow{2}{*}{\textbf{Methods}} & \multicolumn{6}{c|}{\textbf{Metrics}} \\ \cline{3-8}
    \textbf{Testing} & & RMSE $\downarrow$ & REL $\downarrow$ & MAE $\downarrow$ & $\delta_{1.05}\uparrow$ & $\delta_{1.10}\uparrow$ & $\delta_{1.25}\uparrow$ \\ \hline
    \multirow{4}{*}{\begin{tabular}{c}Clear+OOD/\\TransCG\end{tabular}} & \cite{sajjan2020clear} & 0.061 & 0.108 & 0.049 & 33.59 & 54.73 & 92.48 \\
    & \cite{zhu2021rgb} & 0.146 & 0.262 & 0.115 & 13.70 & 26.39 & 57.95 \\
    & \red{\cite{xu2022seeing}} & \red{0.071} & \red{\textbf{0.060}} & \red{0.036} & \red{\textbf{62.99}} & \red{\textbf{82.92}} & \red{95.93} \\
    & Ours & \textbf{0.039} & 0.067 & \textbf{0.030} & 56.68 & 74.61 & \textbf{98.01} \\ \hline
    \multirow{4}{*}{\begin{tabular}{c}TransCG/\\Clear-Real\end{tabular}} & \cite{sajjan2020clear} & 0.085 & 0.095 & 0.052 & 47.26 & 70.76 & 92.54 \\
    & \cite{zhu2021rgb} & 0.152 & 0.225 & 0.139 & 9.86 & 20.63 & 46.02 \\
    & \red{\cite{xu2022seeing}} & \red{0.045} & \red{0.071} & \red{0.040} & \red{33.43} & \red{70.14} & \red{\textbf{99.40}} \\
    & Ours & \textbf{0.041} & \textbf{0.054} & \textbf{0.031} & \textbf{62.74} & \textbf{83.31} & 97.33 \\ \hline
    \end{tabular}
    \end{center}
    \end{minipage}
\end{table}
\begin{table}[t]
    \setlength\tabcolsep{1.5pt}
    \centering
    \begin{minipage}[t]{0.95\linewidth}
    \begin{center}
        \caption{Results of Cross-Domain Experiments of Datasets}
        \label{tab:cross_domain_dataset}
        \begin{tabular}{|c|c|cccccc|} \hline \multirow{2}{*}{\textbf{Training}} &
        \multirow{2}{*}{\textbf{Testing}} & \multicolumn{6}{c|}{\textbf{(Global) Metrics$^*$}} \\ \cline{3-8}
        & & RMSE $\downarrow$ & REL $\downarrow$ & MAE $\downarrow$ & $\delta_{1.05}$ $\uparrow$ & $\delta_{1.10}$ $\uparrow$ & $\delta_{1.25}$ $\uparrow$ \\ \hline
        Clear+OOD & \multirow{2}{*}{TOD} & 0.077 & 0.043 & 0.037 & 83.44 & 89.46 & 95.53 \\
        TransCG & & \textbf{0.044} & \textbf{0.013} & \textbf{0.011} & \textbf{97.54} & \textbf{98.30} & \textbf{98.82} \\ \hline
        Clear+OOD & \multirow{2}{*}{Clear-Real} & \textbf{0.040} & 0.058 & 0.032 & 55.08 & 81.23 & \textbf{98.75} \\
        TransCG & & 0.041 & \textbf{0.054} & \textbf{0.031} & \textbf{62.74} & \textbf{83.32} & 97.33 \\ \hline
        \end{tabular}
    \end{center}
    \vspace{-0.15cm}
    \footnotesize{$^*$: For TOD dataset, global metrics is used since masks are not provided. It denotes that the evaluation is conducted on the full depth map instead of the masked area.}
    \end{minipage}
    \vspace{-0.4cm}
\end{table}
Cross-domain experiments are performed to verify the robustness of our proposed depth completion method and the generalization ability of our proposed TransCG dataset.

For cross-domain experiments of different methods, two experiments are conducted, namely (1) test the performance on our TransCG dataset after training on ClearGrasp synthetic dataset and Omniverse object dataset; (2) test the performance on ClearGrasp real-world dataset after training on our TransCG dataset. Results shown in Tab. \ref{tab:cross_domain} reveal that our method is the most robust one among all methods. Also, it is worth noticing that although LIDF-Refine~\cite{zhu2021rgb} reaches satisfactory results when training domain is similar to the testing domain, the cross-domain testing decreases its performance a lot since its local implicit depth function is environment-dependent. On the contrary, our method is less sensitive to domain changes and is able to achieve satisfactory results under different environment settings.

\red{For cross-domain experiements of different datasets, we select our method as the depth completion model to test the generality of the training dataset on a third-party dataset TOD~\cite{liu2020keypose} and ClearGrasp real-world dataset. Results shown in Tab. \ref{tab:cross_domain_dataset} demonstrate that our real-world dataset is more universal compared to the previous synthetic datasets, even though ClearGrasp real-world dataset has a similar environment as ClearGrasp synthetic dataset which is used for training. The performance differences also reflect the shortcomings of synthetic datasets compared to real-world datasets.}

\subsection{Real Robot Experiments}

\begin{figure}[htbp]
    \vspace{0.4cm}
	\centering
	\includegraphics[width=0.2\textwidth]{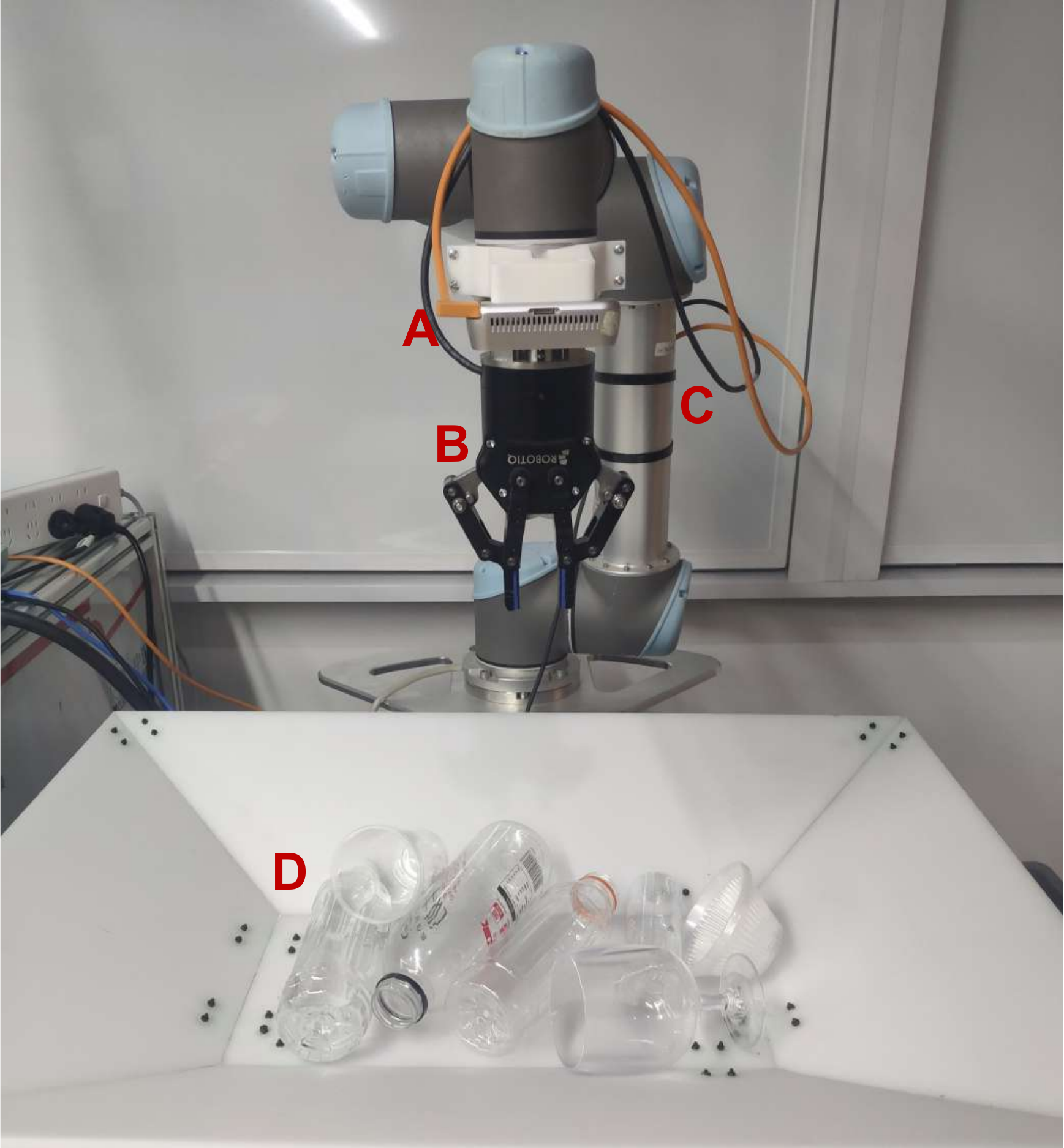}
	\caption{Real robot experiments on a cluttered scene that consists of daily transparent objects. A: Intel RealSense RGB-D camera. B: Robotiq two-finger gripper. C: UR-5 Robot. D: Daily transparent objects.} 
	\label{fig:real-experiment}
	\vspace{-0.2cm}
\end{figure}

To verify the performance of our method in real-world settings, we conduct real robot object grasping experiments. The object grasping pipeline incorporated with our method is introduced in Sec. \ref{subsec:objectgrasping}. The experiments are conducted on a UR-5 robot with an Intel RealSense D435 camera and a Robotiq two-finger gripper, as shown in Fig. \ref{fig:real-experiment}.

We randomly select 8 transparent objects to perform real-robot experiments, 6 of which are completely novel, and the rest 2 objects are the same objects without fixers and markers from our testing set. For every experiment, we randomly put the objects and repeat grasping until an object fails for 3 times. The success rate is defined as $\frac{\# \textrm{objects}}{\# \textrm{attempts}}$, and the completion rate is defined as $\frac{\# \textrm{successfully-grasped objects}}{\# \textrm{objects}}$. Table \ref{tab:real-experiment} reports the experiment results, which shows the effectiveness and feasibility of our method. More details are presented in supplementary video.

\begin{table}[h]
    \setlength\tabcolsep{1.5pt}
    \centering
    \begin{minipage}[t]{0.95\linewidth}
    \begin{center}
    \caption{Results of Real Robot Experiments}
    \label{tab:real-experiment}
    \begin{tabular}{|c|c|c|c|c|} \hline
     & \textbf{\# Objects} & \textbf{\# Attempts} & \textbf{Success Rate} & \textbf{Completion Rate} \\ \hline
    \textbf{Exper. 1} & 7 & 11 & 63.6\% & 100.0\% \\ \hline
    \textbf{Exper. 2} & 7 & 8 & 87.5\% & 100.0\% \\ \hline
    \textbf{Exper. 3} & 7 & 9 & 77.8\% & 100.0\% \\ \hline
    \textbf{Exper. 4} & 8 & 8 & 100.0\% & 100.0\% \\ \hline
    \textbf{Exper. 5} & 8 & 10 & 80.0\% & 100.0\% \\ \hline
    \textbf{Total} & 37 & 46 & 80.4\% & 100.0\% \\ \hline
    \end{tabular}
    \end{center}
    \end{minipage}
    \vspace{-0.5cm}
\end{table} 
\section{Conclusion}\label{sec:conclusion}

In this paper, we propose \textbf{TransCG}, the first large-scale real-world dataset for transparent object depth completion and grasping, built by our novel data collecting pipeline. \red{Our dataset fills the synthetic-to-real gap in the transparent depth completion area and is more general compared to previous synthetic datasets in real-world environments. Moreover, we propose an end-to-end depth completion network \textbf{DFNet}, which is more efficient and robust compared to previous methods according to experiments on various datasets. Real robot experiments of grasping also demonstrate that our method is applicable in real-world settings with novel objects. The compatibility of our method with depth-based manipulation methods allows it to become a default pre-processing step for all downstream tasks concerning transparent objects.}

\printbibliography

\end{document}